\documentclass[sigconf]{acmart}

\AtBeginDocument{%
  }

\setcopyright{rightsretained}
\copyrightyear{2025}
\acmYear{2025}
\acmConference{SIGGRAPH Posters '25}{August 10-14, 2025}{Vancouver, BC, Canada}\acmBooktitle{Special Interest Group on Computer Graphics and Interactive Techniques Conference Posters (SIGGRAPH Posters '25), August 10-14, 2025}\acmDOI{10.1145/3721250.3742976}
\acmISBN{979-8-4007-1549-5/2025/08}

\begin{document}

\title{Co-Speech Gesture and Facial Expression Generation for Non-Photorealistic 3D Characters}

\author{Taisei Omine}
\authornote{Both authors contributed equally to this research.}
\email{Taisei.Omine@sony.com}
\orcid{0000-0001-6558-2046}
\affiliation{%
  \institution{Sony Group Corporation}
  \city{Osaki}
  \state{Tokyo}
  \country{JAPAN}
}

\author{Naoyuki Kawabata}
\authornotemark[1]
\orcid{0009-0004-6329-7371}
\email{Naoyuki.Kawabata@sony.com}
\affiliation{%
  \institution{Sony Group Corporation}
  \city{Osaki}
  \state{Tokyo}
  \country{JAPAN}
}

\author{Fuminori Homma}
\orcid{0009-0009-5002-846X}
\email{Fuminori.Homma@sony.com}
\affiliation{%
  \institution{Sony Group Corporation}
  \city{Osaki}
  \state{Tokyo}
  \country{JAPAN}
}

\begin{abstract}
With the advancement of conversational AI, research on bodily expressions, including gestures and facial expressions, has also progressed. However, many existing studies focus on photorealistic avatars, making them unsuitable for non-photorealistic characters, such as those found in anime. This study proposes methods for expressing emotions, including exaggerated expressions unique to non-photorealistic characters, by utilizing expression data extracted from comics and 
dialogue-specific semantic gestures. A user study demonstrated significant improvements across multiple aspects when compared to existing research.
\end{abstract}
\begin{CCSXML}
<ccs2012>
   <concept>
       <concept_id>10010147.10010371.10010352</concept_id>
       <concept_desc>Computing methodologies~Animation</concept_desc>
       <concept_significance>500</concept_significance>
       </concept>
 </ccs2012>
\end{CCSXML}

\ccsdesc[500]{Computing methodologies~Animation}

\keywords{Facial Expression, NPR, Co-speech Gesture Synthesis, Multi-modality}

%\received{20 February 2025}
%\received[revised]{? ? 2025}
%\received[accepted]{? ? 2025}

% \begin{figure}
%     \centering
%     \includegraphics[width=1\linewidth]{img/fig1_new2.pdf}
%     \caption{Exaggerated Expressions Targeted in This Study}
%     \label{fig:manpu}
% \end{figure}
% % \begin{teaserfigure}
% \begin{figure}
%   \includegraphics[width=\textwidth]{img/fig1_new2.pdf}
%   \caption{Exaggerated Expressions Targeted in This Study}
%   % \Description{Enjoying the baseball game from the third-base
%   % seats. Ichiro Suzuki preparing to bat.}
%   \label{fig:manpu}
% % \end{teaserfigure}
% \end{figure}

% \begin{figure}
\begin{teaserfigure}
    \centering
    \includegraphics[width=1\linewidth]{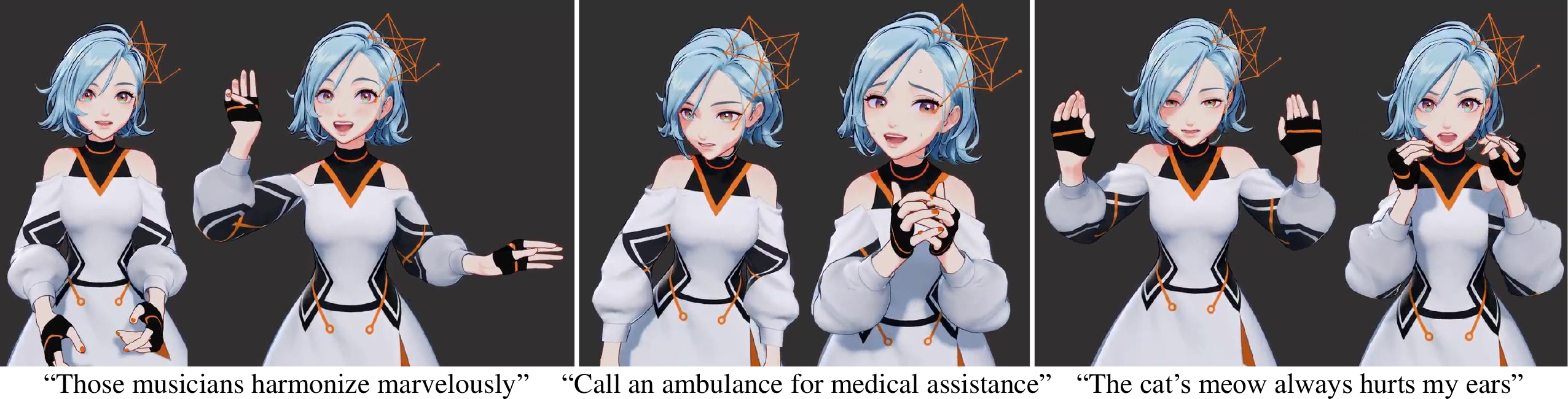}
    % \caption{Examples of the Results and Transcripts Presented in the User Study, Comparing EMAGE (Left) and Our Method (Right).}
    \caption{Examples of the Results and Transcripts Presented in the User Study, Comparing EMAGE (Left) and Ours (Right).}
    \Description{Three examples comparing Emage and Ours. Each example shows an image with Emage's results applied to an avatar on the left, and an image with Ours' results applied on the right. Below each image, the input dialogue text is displayed. The texts are: 'Those musicians harmonize marvelously', 'Call an ambulance for medical assistance', and 'The cat's meow always hurts my ears'.}
    \label{fig:example_of_videos}
% \end{figure}
\end{teaserfigure}

\maketitle

\section{Introduction}
% 対話AIの発展に伴い、ジェスチャーや表情といった身体表現の獲得も期待されている。
% しかしながら、ジェスチャや表情の研究は、多くがフォトリアルなデジタルヒューマンなどを対象としているため\cite{liu2024emage}、
% ノンフォトリアルキャラクターに最適なジェスチャ・表情・誇張表現を実現できない。
% また、既存データセットは対話に特化していないことが多く、より適切なデータセットの開発が必要であると考えられる。
% 本研究では、漫画から表情データを自動抽出し、また、キャラクター対話に特化したセマンティックなジェスチャデータを収集・合成することで、特にアニメ調の3Dのノンフォトリアルキャラクターとの対話に特化したジェスチャおよび表情のセリフに合わせた制御手法を提案する。
With the advancements in conversational AI, there is increasing interest in integrating bodily expressions. 
% of non photorealistic character
% However, most of the existing research focuses on photorealistic digital humans \cite{liu2024emage}, limiting the applicability of these approaches to non-photorealistic characters like those found in anime. This limitation arises because the naturalness of gestures and expressions differs significantly between non-photorealistic and photorealistic entities. Additionally, current methods are inadequate for capturing unique non-photorealistic expressions, including exaggerated gestures. Moreover, many existing techniques are not specifically tailored for dialogue interactions, highlighting the need for more appropriate training datasets for such applications. 
However, most research on gesture and facial expression has focused on photorealistic digital humans \cite{liu2024emage}, and thus it fails to achieve optimal gesture, facial expressions, and exaggerated expressions for non-photorealistic characters. 
% Moreover, existing datasets are often not specifically designed for dialogue scenarios, as most of them are retrieved from speech data.
Additionally, existing datasets are rarely designed specifically for dialogue scenarios, being primarily derived from speech data.
% highlighting the need for the development of more suitable datasets.
% In this study, we propose a control method designed to align gestures and expressions with dialogues for non-photorealistic 3D characters in an anime style. 
% This method involves the automatic extraction of expression data from comics, as well as the collection and synthesis of semantic gesture data specialized for character dialogues.
% We propose a method that aligns gestures and expressions with dialogues for non-photorealistic 3D characters by collecting and synthesizing semantic gestures for character dialogues and automatically extracts expression data from comics.
We propose a method that aligns gestures and expressions with dialogues for non-photorealistic 3D characters. Our approach collects and synthesizes semantic gestures for character dialogues and automatically extracts expression data from comics.

\section{Method}
% ノンフォトリアルキャラクターの制御は、ジェスチャ・表情で別々の制御モデルを用いて統合する。これにより、それぞれがより適したデータソースからのデータの調達が可能となる。
%削除Our proposed method is achieved by using separate methods for gestures and facial expressions. 
%削除This approach allows for the collection of data from the most appropriate sources for each component.
% ノンフォトリアルキャラクターの制御は、ジェスチャ・表情で別々の制御モデルを用いて統合する。これにより、それぞれがより適したデータソースからのデータの調達が可能となる。%になり、アニメ調のノンフォトリアルキャラクターに合った自然かつ多様な振る舞いが可能になる。
%短くした Our method separates gesture and facial expression, enabling data collection from the most appropriate sources for each.

\subsection{Gesture}
% Traditional 
Prior gesture generation research has focused mainly on photorealistic avatars, 
%置き換え aiming to generate realistic movements by incorporating semantic information to audio-based data \cite{liu2024emage}.
using semantic and audio data to create realistic movements \cite{liu2024emage}.
%削除 , Zhang2024SemanticGesture
%置き換え However, as the movements of realistic humans differ from those required for non-photorealistic characters, conventional techniques are considered unsuitable for application to non-photorealistic characters. 
%置き換え Non-photorealistic characters require movements that prioritize clarity over realism. 
However, conventional techniques designed for realistic human movements are unsuitable for non-photorealistic characters that prioritize clarity over realism. To address this, we propose a semantic-first motion generation method.

\subsubsection{Gesture Dataset for Conversational Character}
%置き換え In proposing a gesture generation method for non-photorealistic characters, we prepared a new gesture dataset specifically designed for dialogue interactions.
For our non-photorealistic character gesture generation method, we created a new dialogue-focused gesture dataset of 500 instances.
%置き換え Many of the existing datasets \cite{liu2022beat} are recorded in a style where a single speaker continues to speak unilaterally, as in speeches or presentations. We determine this is insufficient for realizing effective character AI interaction.
%置き換え Our newly created gesture dataset comprises 500 instances developed using motion capture recordings and refined through animator-led motion cleanup.
Unlike existing datasets \cite{liu2022beat} that feature unilateral speech patterns from presentations or speeches, our motion-captured dataset focuses on interactive dialogue scenarios to better support conversational character AI.
% These gesture data are
%置き換え Each gesture data sample is linked to phrases commonly used in dialogues. These phrases are composed of words used in various contexts such as greetings, emotional expressions, emphasis, iconic gestures, active listening, and gaze guidance.Utilizing this dataset makes character dialogue more interactive and engaging.
Each gesture sample links to common dialogue phrases spanning contexts like greetings, emotions, emphasis, iconic gestures, active listening, and gaze guidance, making character interactions more engaging.

\subsubsection{Gesture Synthesis based on Semantic Phrase Retrieval}
% The input for the proposed method is the text of the lines intended for the character to speak. 
%置き換えThe proposed method uses the character’s dialogue as input. The input text is divided into phrases and converted into text embeddings. 
Our method processes character dialogue by dividing input text into phrases and converting them to embeddings.
%置き換えThese embeddings are then compared with the embeddings of the phrases in the gesture dataset to identify the most similar phrase. If no similar phrase is found, a neutral dialogue gesture is randomly selected.
These are compared with gesture dataset phrase embeddings to find similarities, defaulting to random neutral gestures when no match exists.
%置き換えBy integrating and smoothing the gesture data linked to similar phrases, we generate gesture sequence data that matches the entire input. 
%置き換えSince this gesture sequence data is generated solely from textual input, it is not synchronized with audio at this stage. However, synchronization can be achieved by adjusting the gesture sequence lengths to align with the duration of the character's spoken audio.
We generate entire gesture sequences by integrating and smoothing matched gesture data, then adjust them to align with the character's speech duration.
%置き換え In the evaluation experiment described later, the phrase data linked to gestures is composed in Japanese. We employ SentenceBERT\footnote{https://huggingface.co/sonoisa/sentence-bert-base-ja-mean-tokens-v2/tree/main}, a text encoder trained on Japanese, for text embedding. Cosine similarity between embeddings is utilized for similarity calculation. For phrase division, punctuation marks in the Japanese lines are detected, and sentences are divided into phrases accordingly. When English sentences are input, they are first translated into Japanese before being processed by the system.
For our evaluation experiment, we use Japanese phrase data linked to gestures and employ SentenceBERT\footnote{https://huggingface.co/sonoisa/sentence-bert-base-ja-mean-tokens-v2/tree/main} for text embedding, with cosine similarity for matching. 
% Phrases are divided at Japanese punctuation marks. 
English inputs are first translated to Japanese before processing.

%%%%%%%%%%%%%%%%%%%%%%%%%%%%%%%%%%%%%%%%%%%%%%%%%%%%%%%%%%%%%%%%%%%%%%%%%%%%%%%%%%%%%%%%%%%%
\subsection{Facial Expression Generation}
\subsubsection{Dataset Development}

\begin{figure}
    \centering
    \includegraphics[width=1\linewidth]{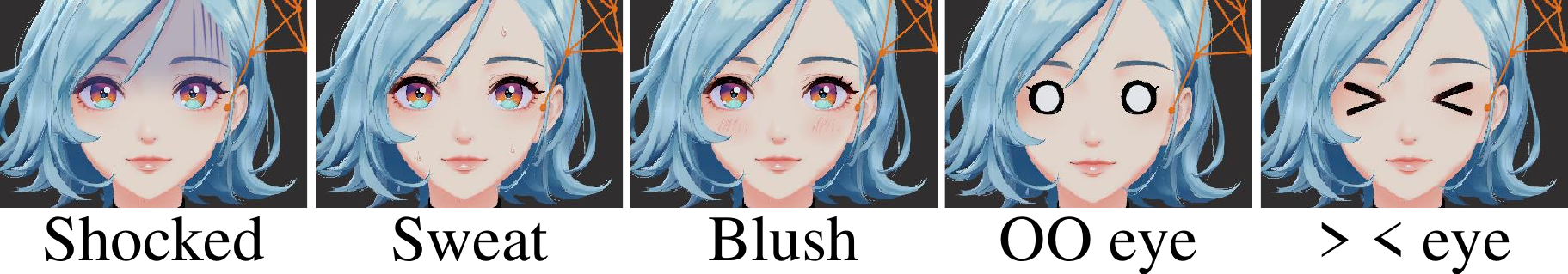}
    \caption{Exaggerated Expressions Targeted in This Study}
    \Description{Five anime-style avatar faces are arranged in a row, each with a different expression. From left to right: a shocked face with blue lines on the forehead, a sweating face, a blushing face, a cartoon-style face with eyes drawn only as circles, and a face with eyes shaped like inequality symbols(><).}
    \label{fig:manpu}
\end{figure}

%置き換え Since existing photorealistic facial expression datasets were unsuitable for non-photorealistic characters, we developed a novel dataset. This dataset also targets exaggerated expressions, as illustrated in Fig. \ref{fig:manpu}.
We developed a novel dataset for non-photorealistic characters including exaggerated expressions (Fig. \ref{fig:manpu}), as existing datasets were designed for photorealistic humans.
% 我々は、漫画から表情を抽出し、3Dキャラクターに適用するためのブレンドシェイプ値に変換した。漫画を用いることで、誇張表現を含む様々な表情を抽出できる。漫画データは、Manga109\cite{multimedia_aizawa_2020}を利用した。
% データセットから切り出したキャラクターの顔部分の画像を入力し、次の３つの手法で表情を推測することで、効率的に3Dキャラクターのブレンドシェイプ値に変換する。
%置き換え We extracted facial expressions from comics and converted them into blendshape values for application to 3D characters. Several blendshape values correspond to exaggerated expressions. Utilizing comics allows us to capture a wide range of non-photorealistic expressions, including exaggerated ones. 
We extracted facial expressions from comics and converted them to 3D character blendshape values, including several for exaggerated expressions. This comic-based approach captures a wide range of non-photorealistic expressions.
% For the comic data, we employed the Manga109 dataset \cite{multimedia_aizawa_2020}. 
% Using the cropped character face images from the dataset, we inferred facial expressions with the following three methods and efficiently converted them into blendshape values.
Specifically, we used cropped character face images from a dataset \cite{multimedia_aizawa_2020} as input. We inferred facial expressions using the following three methods:% and then efficiently converted them into blendshape values: 
% 1. Extraction of tags related to facial expressions and exaggerated features using the Stable Diffusion Tagger extension \footnote{https://github.com/picobyte/stable-diffusion-webui-wd14-tagger}.
% 2. Prediction of facial feature positions and shapes using facial landmarks extracted from the Anime Face Detector\footnote{https://github.com/hysts/anime-face-detector}.
% 3. Extraction of additional information through multiple-choice answers using image-enabled LLM\cite{openai2024gpt4ocard}. Incorporating dialogue and context into the prompts further enhanced the accuracy of this method.
(1) extraction of tags related to facial expressions and exaggerated features using the Stable Diffusion Tagger\footnote{https://github.com/picobyte/stable-diffusion-webui-wd14-tagger}; 
(2) prediction of facial feature positions and shapes using facial landmarks detected by the Anime Face Detector\footnote{https://github.com/hysts/anime-face-detector}; 
and (3) extraction of additional information through multiple-choice answers using 
% an image-enabled 
a multimodal 
LLM
% \cite{openai2024gpt4ocard}
, with character's dialogue and context incorporated to further enhance accuracy.
% これらの手法による出力をもとに、最終的な表情パラメータ値をルールベースで決定する。
% 最終的に10,000以上のデータを自動的に作成した。
From the outputs of these three methods, we derived the final blendshape values using a rule-based approach, ultimately generating over 10,000 data points automatically.
% 特に誇張表現については、Shockedが0.○○\%、Sweatが○○\%、...含まれる。
% およそ50\%のデータポイントにいずれかの誇張表現が含まれる。
% 50\% of the data points contain exaggerated expressions.
% Approximately 50\% of the data points contain one or more of the exaggerated expressions shown in Fig. \ref{fig:manpu}.
Approximately 50\% of the data points contain one or more of the exaggerated expressions shown in Fig. \ref{fig:manpu}, though some are infrequent in comics and thus significantly underrepresented in the dataset.

\subsubsection{Similar Face Retrieval}
% 次に、作成した表情データをもとに、どの表情を使うかをセリフを元に出力するシステムを作成した。以下の手順で最適な表情データが選択される。
Next, based on the 
%削除generated facial expression 
dataset, we developed a system that determines which data point to use based on an avatar's dialogue. The data point selection process consists of two steps:
% \begin{enumerate}
    % \item \textbf{表情データへの感情値の紐づけ} 顔画像やセリフなどをもとに、表情データに紐づく感情値を画像対応のLLM\cite{openai2024gpt4ocard}を用いて設定する。具体的には、130種類程度の感情の中から最も当てはまる数種類の感情とその強度を推定する。出力例として、\{喜び: 0.8, 楽しさ: 0.7, 興味: 0.65\} があり、これをベクトルに変換した。
    %削除\item \textbf{Linking Emotion Values to Data Points} 
(1) We assign emotion values to every data point by utilizing dialogues and facial images with 
% an image-compatible 
a multimodal 
LLM. Specifically, we estimate several relevant emotions and their intensities from approximately 130 emotion categories. For example, an output might be \{Joy: 0.8, Amusement: 0.7, Interest: 0.65\}.
    %削除, which is then converted into a vector representation.
    % \item \textbf{発話セリフの類似度判定} 発話セリフについてもセリフから感情値を推論する。その感情値と最もcosine類似度が高い感情値をもつ表情データをキャラクターに適用する。
    %削除\item \textbf{Determining Similarity of Avatar's Dialogue} 
(2) We infer emotion values from an avatar's dialogue using the same LLM and apply the data point that has the highest cosine similarity to these inferred emotion values.
% \end{enumerate}
% これにより、任意のセリフに対して、そのセリフにあったの表情を適用することができる。実際の利用時には、表情のトランジション機能や、発話音声から唇の形を推測するリップシンク機能、まばたきの機能を追加して、動きをつけている。
This method enables the application of facial expressions that are appropriate for any 
% given 
dialogue. In practice, 
we enhance the system with facial expression transitions, lip-syncing
%削除that infers lip shapes from TTS audio
, and blinking to create a more natural appearance. 
% なお、リップシンク機能は表情を極力阻害しないよう調整される。
%削除Note that the lip-sync function is adjusted to minimize interference with facial expressions.

\section{Evaluation and Conclusion}

\begin{table}%[htbp]
    % \caption{Combined Percentage of "Agree" and "Somewhat Agree" Responses in the User Study}
    \caption{Sum of "Agree" and "Somewhat Agree" Response Percentages in the User Study}
    \centering
    \begin{tabular}{@{}l|@{}c@{}c@{}c@{}c@{}c}
    \hline
    % & EMAGE & &&&&Ours \\
    & \multicolumn{5}{c}{EMAGE / Ours} \\
    \cline{2-6}
    \multicolumn{1}{c|}{\%}
    & {\begin{tabular}{c}Overall\\Appeal\end{tabular}} 
    & {\begin{tabular}{c}Text\\Reflection\end{tabular}} 
    & {\begin{tabular}{c}Visual\\Match\end{tabular}} 
    & {\begin{tabular}{c}Audio\\Sync\end{tabular}} 
    & {Diversity} \\
    % & {\begin{tabular}{c}Overall\\Appeal\end{tabular}} 
    % & {\begin{tabular}{c}Text\\Reflection\end{tabular}} 
    % & {\begin{tabular}{c}Visual\\Match\end{tabular}} 
    % & {\begin{tabular}{c}Audio\\Sync\end{tabular}}
    % & {Diversity} \\
    \hline
    Angry & 11 / \textbf{50} &29 / \textbf{78} & 12 / \textbf{50} &20 / \textbf{76} & 32 / \textbf{84} \\
    % Disgusted & 9.6 / \textbf{49.6}& 30.4 / \textbf{48.0} & 8.0 / \textbf{43.2}& 27.2 / \textbf{74.4}& 28.0 / \textbf{76.0}  \\
    Disgusted & 10 / \textbf{50}& 30 / \textbf{48} & 8 / \textbf{43}& 27 / \textbf{74}& 28 / \textbf{76}  \\
    % Contempt & 8.8 / \textbf{55.2} & 24.8 / \textbf{48.0}& 8.8 / \textbf{66.4}& 25.6 / \textbf{72.8} & 16.0 / \textbf{56.0} \\
    Contempt & 9 / \textbf{55} & 25 / \textbf{48}& 9 / \textbf{66}& 26 / \textbf{73} & 16 / \textbf{56} \\
    % Fear & 23.2 / \textbf{66.4}& 40.8 / \textbf{72.8}& 21.6 / \textbf{70.4} & 28.8 /  \textbf{80.0}& 48.0 / \textbf{92.0} \\
    Fear & 23 / \textbf{66}& 41 / \textbf{73}& 22 / \textbf{70} & 29 /  \textbf{80}& 48 / \textbf{92} \\
    % Happy & 20.0 / \textbf{87.2}& 28.0 / \textbf{88.0}& 24.0 / \textbf{88.0} & 36.8 / \textbf{88.0}& 32.0 / \textbf{96.0} \\
    Happy & 20 / \textbf{87}& 28 / \textbf{88}& 24 / \textbf{88} & 37 / \textbf{88}& 32 / \textbf{96} \\
    % Sad & 12.0 / \textbf{62.4}& 28.0 / \textbf{67.2}& 9.6 / \textbf{67.2} & 32.0 / \textbf{72.8}& 32.0 / \textbf{88.0}\\
    Sad & 12 / \textbf{62}& 28 / \textbf{67}& 10 / \textbf{67} & 32 / \textbf{73}& 32 / \textbf{88}\\
    % Surprised & 28 / \textbf{53}& 33 / \textbf{58}& 21 / \textbf{58}& 38 / \textbf{71}& 36 / \textbf{68} \\
    Surprised & 28 / \textbf{53}& 33 / \textbf{58}& 21 / \textbf{58}& 38 / \textbf{71}& 36 / \textbf{68} \\
    \cline{1-6}
    % Avg. & 16.1 / \textbf{60.5}& 30.5 / \textbf{65.6}& 15.0 / \textbf{63.3} & 29.8 / \textbf{76.5} & 32.0 / \textbf{80.0} \\
    Avg. & 16 / \textbf{61}& 31 / \textbf{66}& 15 / \textbf{63} & 30 / \textbf{77} & 32 / \textbf{80} \\
    \hline
    \end{tabular}
    \label{tab:eval_result}
\end{table}

% ノンフォトリアルなキャラクターに特化した評価は、先行研究\cite{liu2024emage}の評価指標による比較が難しかったため、本論ではユーザースタディにより評価を行った。
% Since it was challenging to compare the evaluation of non-photorealistic characters using the metrics from prior work \cite{liu2024emage}, we evaluated them through a user study.
%削除As comparing non-photorealistic character results with metrics from prior work \cite{liu2024emage} presented challenges, we instead evaluated through a user study.
% 評価は、8つの感情ごとに異なるセリフとその読み上げ音声を含む、MEADデータセット\cite{kaisiyuan2020mead}を用いた。その中からNeutralを除いた7種の感情ごとに、5個のセリフをランダムに選択し、計35個のデータを用いた。音声は、最も感情的であるlevel3の音声を用いた。
%書き換えThe evaluation employed the MEAD dataset \cite{kaisiyuan2020mead}, which contains distinct utterances and corresponding speech recordings for eight emotions. From this dataset, we selected five utterances for each of the seven emotions excluding Neutral, resulting in a total of 35 samples. 
Evaluation was conducted through a user study using 35 samples selected from the MEAD dataset \cite{kaisiyuan2020mead}, which includes distinct utterances and corresponding speech recordings for eight emotions.
%削除The level 3 audio recordings, representing the highest intensity of emotion, were used.  
% 比較する先行研究は、表情とジェスチャを生成可能な研究である、EMAGE\cite{liu2024emage}を用いた。これは、音声を元に、フォトリアルな人間を再現した表情とジェスチャを出力する。
For comparison, we used EMAGE \cite{liu2024emage}, a study capable of generating both photorealistic facial expressions and gestures from speech audio. 
% 被験者は、我々とEMAGEの出力をリファレンスキャラクターに当てはめて作成された計70個の動画を閲覧し、質問に対して"当てはまらない"から"当てはまる"の5段階評価を行った。
%追記
25
participants viewed a total of 70 videos created by mapping the outputs of either our approach or EMAGE onto a reference character. Each video was rated on a five-point Likert scale ranging from “not applicable” to “applicable.”
% 質問は、各動画ごとに、"全体的な印象として魅力的である(Overall Appeal)", "発話テキストの内容や感情を反映している(Text Reflection)", "キャラクターの見た目と合っている(Appearance Match)", "音声と、動き・表情の同期がとれている(Audio Sync)"の4つある。
% Four questions were posed for each video: Overall Appeal (whether the overall impression is appealing), Text Reflection (whether the content and emotion of the spoken text are conveyed), Visual Match (whether it matches the character’s appearance), Audio Sync (whether the synchronization between audio and gestures/facial expressions is accurate)  
% For each video, four questions were asked: Overall Appeal, which determines how appealing the video is overall; Text Reflection, which assesses whether the content and emotion of the spoken text are successfully conveyed; Visual Match, which evaluates how well the gestures or facial expressions align with the character’s appearance; and Audio Sync, which judges the precision of synchronization between audio and the gestures or facial expressions.
For each video, four questions were asked: Overall Appeal (general appeal), Text Reflection (reflection of spoken text’s content and emotion), Visual Match (alignment of gestures or facial expressions with character appearance), and Audio Sync (accuracy of audio synchronization with gestures/facial expressions).  
% また、感情ごとに5つの動画を見た後に、"動きや表情は多様である(Diversity)"かを5段階で選択する質問がある。
% In addition, after watching five videos per emotion, participants rated Diversity (the diversity of the gestures and facial expressions) on a five-point scale. 
In addition, after watching five videos per emotion, participants rated Diversity (diversity of gestures/facial expressions) on the same scale.
% 実験は25人の主に日本人の被験者で行い、セリフは英語および、日本語訳を提示した。
Fig. \ref{fig:example_of_videos} shows examples of the results and transcripts presented in the user study.
%削除The experiment involved 25 participants.
%削除, predominantly Japanese speakers, and both the original English utterances and their Japanese translations were provided to ensure comprehension.
% Table \ref{tab:eval_result}は、ユーザースタディにおいて、”当てはまる”または”やや当てはまる”と答えた割合の合計を示している。フォトリアル向けであるEMAGEに比べ、全項目の評価が向上していることがわかる。
%置き換えTable \ref{tab:eval_result} presents the cumulative percentage of instances where responses were either “applicable” or “somewhat applicable”. 
Table \ref{tab:eval_result} shows the combined percentage of "applicable" and "somewhat applicable" responses.
The results indicate significant improvement across all evaluation criteria
compared to EMAGE.
%削除, which is designed for photorealistic outputs
%削除
% \section{Conclusion}
% % 本論では、ノンフォトリアルキャラクターに適合したジェスチャと表情、誇張表現を表現するためのシステムを提案した。
% In this paper, we proposed a method for expressing gestures, facial expressions, and exaggerated expressions optimized for non-photorealistic characters.
% % また、対話に特化したジェスチャー生成を提案した。
% %削除Furthermore, we introduced a gesture generation method for conversational interactions.
% % これらの効果として、提案手法を用いることで、ノンフォトリアルキャラクターに、最新のフォトリアル向けのシステムを組み込んだ場合と比較して、ユーザーの満足度は大きく向上する可能性を示した。
% % これらの効果として、, compared to incorporating the latest photo-realistic system into a non-photorealistic character, our approach showed the potential to significantly increase user satisfaction.
% As a result, our proposed approach demonstrated the potential to significantly enhance user satisfaction compared to integrating state-of-the-art systems designed for photorealistic characters into a non-photorealistic character.
% % 今後は、さらに多様な見た目や性格のキャラクターに適用できるよう改善を進める。
We plan to further enhance the system to accommodate characters with a greater variety of appearances and personalities.

\bibliographystyle{ACM-Reference-Format}
\bibliography{references}

\end{document}